\documentclass{article}

\usepackage[preprint,nonatbib]{neurips_2024}

\usepackage{graphicx}
\usepackage[utf8]{inputenc} 
\usepackage[T1]{fontenc}    
\usepackage{hyperref}       
\usepackage{url}            
\usepackage{booktabs}       
\usepackage{amsfonts}       
\usepackage{nicefrac}       
\usepackage{microtype}      
\usepackage{xcolor}         
\usepackage{graphics}
\usepackage{graphicx}
\usepackage{tabularx}
\usepackage{subfig}
\usepackage{amsmath}

\title{Machine Learning for the Digital Typhoon Dataset: Extensions to Multiple Basins and New Developments in Representations and Tasks}

\author{%
Asanobu Kitamoto$^{1,2}$\quad
Erwan Dzik$^{3,1}$\quad
Gaspar Faure$^{4,1}$\\
  $^{1}$ National Institute of Informatics, Japan \\
  $^{2}$ Typhoon Science and Technology Research Center, Yokohama National University, Japan \\
  $^{3}$ Institut National Polytechnique de Grenoble, France\\
  $^{4}$ Polytechnique Montréal, Canada\\
}

\begin{document}

\maketitle

\begin{abstract}
  This paper presents the Digital Typhoon Dataset V2, a new version of the longest typhoon satellite image dataset for 40+ years aimed at benchmarking machine learning models for long-term spatio-temporal data. The new addition in Dataset V2 is tropical cyclone data from the southern hemisphere, in addition to the northern hemisphere data in Dataset V1. Having data from two hemispheres allows us to ask new research questions about regional differences across basins and hemispheres. We also discuss new developments in representations and tasks of the dataset. We first introduce a self-supervised learning framework for representation learning. Combined with the LSTM model, we discuss performance on intensity forecasting and extra-tropical transition forecasting tasks. We then propose new tasks, such as the typhoon center estimation task. We show that an object detection-based model performs better for stronger typhoons. Finally, we study how machine learning models can generalize across basins and hemispheres, by training the model on the northern hemisphere data and testing it on the southern hemisphere data. The dataset is publicly available at \url{http://agora.ex.nii.ac.jp/digital-typhoon/dataset/} and \url{https://github.com/kitamoto-lab/digital-typhoon/}.
\end{abstract}

\section{Introduction}

Tropical cyclones, also known as typhoons and hurricanes in certain regions, have been the critical research target due to their substantial societal impact \cite{emanuel_100_2018}. In addition to traditional numerical approaches to studying tropical cyclones, such as atmospheric simulation models, new data-driven approaches such as machine learning approaches, have proven to be a promising choice for analyzing or forecasting tropical cyclones. To promote the application of machine learning to tropical cyclones, in November 2023, we released the first version of the Digital Typhoon Dataset \cite{neurips23}.

This dataset bridged the gap between meteorology and machine learning communities by providing the standard dataset for tropical cyclones. This dataset is the longest satellite typhoon image dataset, spanning more than 40 years from 1978 to 2022, consisting of typhoon images created from the meteorological satellite series 'Himawari' and the best track from the Japan Meteorological Agency. After the first version was released, we started preparing the second version as an annual regular update by adding new data from the previous year. We also review the data processing pipeline to respond to user's needs. 

The new dataset, Digital Typhoon Dataset V2, however, is not just a regular update. The most impactful addition is the data from the southern hemisphere, which allows us to ask new research questions: How can the model learned in one basin be transferred to another region or hemisphere? We expect that comparison between hemispheres and basins may lead to new data-driven discoveries in the regional characteristics of tropical cyclones. This dataset will be introduced in Section \ref{sec-dataset}. 

In addition to the dataset release, this paper also reports our recent developments in data representations. Section \ref{sec-rep} describes the use of self-supervised learning (SSL) framework for representation learning (MoCo v2) applied to typhoon intensity forecasting and extra-tropical storm transition forecasting. Due to the high dimensionality and temporal dependency of the data, we show there are opportunities to improve representation learning. 

We will also discuss new tasks. In Section \ref{CET}, we will introduce the typhoon center estimation task; namely to identify the center of the typhoon only from satellite images. This is an operationally critical task because many downstream tasks, such as analysis and forecasting tasks in Section \ref{sec-rep}, depend on the estimated center of the typhoon. This task is not difficult when the eye is visible for a strong typhoon, but is challenging when the cloud pattern is irregular for a weak typhoon. We propose an object detection-based method to estimate the center from a heatmap image obtained by a U-Net architecture. 

In Section \ref{sec-basin}, we will also discuss how a machine learning model can transfer or generalize across basins and hemispheres. Meteorologically speaking, tropical cyclones in different basins are the same meteorological phenomena, but subtle regional differences may be found by data-driven methods. While we study how a machine learning model trained in one basin can transfer well to another basin, we also need to consider data quality issues and interagency discrepancies of the best track dataset across basins. 

\section{Digital Typhoon Dataset V2}
\label{sec-dataset}

\subsection{Updates on Temporal and Spatial Dimensions}

The extension to the temporal dimension is a regular annual update after the best track data for the previous typhoon season is released from the Japan Meteorological Agency (JMA). The typhoon season starts in January and ends in December, and the best track data for the previous typhoon season is typically released in February. Based on the updated best track data, we added the data for the 2023 typhoon season. The number of tropical cyclones (TC) increased from 1,099 to 1,116, and the number of images increased from 189,364 to 192,956, as summarized in Table \ref{tab:two-basins}. 

On the other hand, the extension to the spatial dimension, the southern hemisphere data based on the best track data from the Bureau of Meteorology (BoM), Australia, is a new addition to the dataset. Because the Himawari satellite observes both the northern and southern hemispheres in one disk image, we can apply the same data processing pipeline to both hemispheres. 

\begin{table}[t]
    \caption{Characteristics of the WP and AU datasets.}
    \label{tab:two-basins}
    \centering
    \begin{tabularx}{\textwidth}{lXX} 
    \toprule
               & WP dataset  &  AU dataset \\ 
      \midrule
     Temporal coverage & 1978-2023 & 1979-2024 \\
     Spatial coverage & 100E to 180E in the Northern Hemisphere  
     & 90E to 160E in the Southern Hemisphere  \\
     Season & January to December & July (the previous year) to June \\
     Best track & Japan Meteorological Agency (JMA) & Bureau of Meteorology (BoM), Australia \\
     Number of TCs & 1,116 & 480 \\
     Number of Images & 192,956 & 70,087 \\
           \bottomrule
    \end{tabularx}
\end{table}

We call the northern hemisphere dataset WP (Western Pacific) and the southern hemisphere dataset AU (Around Australia) to differentiate the two regions. This definition is different from a typical definition of tropical cyclone basins.
For example, IBTrACS \cite{knapp_international_2010-1} separates the Southern Pacific (SP) and Southern Indian (SI),  as shown in the Digital Typhoon website \url{http://agora.ex.nii.ac.jp/digital-typhoon/ibtracs/}. In our dataset, the AU basin combines SP and the eastern part of SI, covering tropical cyclone tracks south of the equator between 90E and 160E. Table \ref{tab:two-basins} summarizes the difference between the WP and AU datasets.  

Important differences between the two hemispheres can be summarized as follows. 
\begin{enumerate}
\item A tropical cyclone (TC) is a typhoon in WP but a cyclone in AU. We should use the tropical cyclone as a general term, but in this paper, we sometimes use the term typhoon to mean both the typhoon and the cyclone. 
\item The TC season is between January and December in WP but between July (the previous year) and June in AU, which means that a year is different from a TC season in the southern hemisphere. Therefore, it is more accurate to use a season instead of a year to mean a 12-month interval.   
\item The direction of circulation is opposite due to the Coriolis effect on the Earth's surface. This difference may significantly impact some machine learning tasks across hemispheres. A typical solution is to flip the image horizontally for one hemisphere so that the direction of circulation is the same for both hemispheres.
\item JMA and BoM both use the Dvorak method for estimating typhoon intensity, but the details of the method differ in many aspects. Comparison across basins and hemispheres should consider interagency differences. The definition of intensity scale (grade) may also be different across agencies. 
\end{enumerate}

To have a uniform definition of values across two datasets, we modified the best track data from the BoM in the following points.

\begin{enumerate}
\item The wind speed is converted from meter per second (m/s) to knot (kt) using the equation $\mathrm{kt} = \frac{\mathrm{m/s}}{0.5144}$.
\item We assign the grade by applying the JMA thresholds of maximum wind speed to the maximum wind speed of the BoM best track. We did not use the type assigned by BoM. 
\end{enumerate}

The dataset format has not been changed from V1 to V2, so Python-based software library \texttt{pyphoon2} can work with Dataset V2 as well as Dataset V1. This library comes with a data loader and components to help build machine learning pipelines. \texttt{pyphoon2}  
is downloadable from \url{https://github.com/kitamoto-lab/digital-typhoon/}.

\subsection{Updates on the Data Processing Pipeline}

The data processing pipeline has one relevant change. The map projection has changed from {\em Lambert azimuthal equal-area projection} to {\em azimuthal equidistant projection}. This is in response to a typical usage of the dataset in meteorology community, such as measuring the diameter of the typhoon eye. To measure the distance between the typhoon center and the target pixel using the Euclidean distance of pixel difference, the ideal map projection should preserve the distance from the center to the pixel (equidistant), rather than preserve the area around the center (equal-area). To accommodate this need for meteorological research, we decided to change the map projection method in the data processing pipeline. Due to this change, the entire dataset was refreshed, so the Dataset V1 cannot be reused as a part of Dataset V2. 

We expect, however, that this change does not make a significant impact on machine learning tasks. First, the two map projections do not exhibit a major difference within the scope (cropping area) of the dataset, namely 1250km from the center. Second, machine learning models can adapt to minor differences in the dataset while training, so the performance of the models is expected to be roughly the same. 

The WP and AU datasets use the same data processing pipeline. The only difference is the source of the best track data: JMA for WP and BoM for AU. Because the satellite data is the same, the temporal and spatial resolution of the dataset is the same. However, note that the frequency of corrupted or missing data is higher in the southern hemisphere, because the southern hemispheres received fewer observations due to sensor problems, especially during Himawari-3, Himawari-4, Himawari-5, and GOES-9. 

Both datasets also use the same interpolation method for the best track. The persistence method is used for the wind speed: the last known wind speed value is kept for every image until the next ground truth value. A linear interpolation is used for the central pressure to fill in the missing data between 2 ground truth values. In total, 152,034 images, representing 78.8\% of the whole dataset, have interpolated metadata. They are considered ground truth in the following machine learning tasks. 

\subsection{Notes on Data Augmentation}
\label{sec-augmentation}

Data augmentation is typically used in machine learning to increase the dataset size. For example, when we have a dataset of natural images, we can flip an image to increase the variation of the dataset because a flipped dog is still a dog, even when flipped upside down. This is possible because the augmentation does not impact the semantic meaning of the image. However, the story is different for typhoon images. Because the direction of circulation is determined by the physical law of nature, a flipping of a typhoon image generates a physically impossible image, which has no value in adding to the dataset. From a meteorologist's point of view, even a tiny rotation may cause physically unrealistic images because the atmosphere reflects a subtle balance between the north (up) and south (down). Hence, data augmentation for the Digital Typhoon Dataset should be designed so as not to break the physical meaning of the image. 

Note that we intentionally use image flipping in Section \ref{sec-basin} to align the direction of circulation across basins. Here the purpose is not data augmentation but data preprocessing, and the image flipping is applied to all the datasets. 

\subsection{Dataset Availability}

The Digital Typhoon Dataset V2 is available at the official page \url{http://agora.ex.nii.ac.jp/digital-typhoon/dataset/} with an open data license, namely the Creative Commons Attribution 4.0 International (CC BY 4.0) License. The dataset V1 is preserved as obsolete and may be removed in future updates. The following is a recommended data citation \cite{dt-dataset}. 

\begin{quote}
    Digital Typhoon Dataset V2 (National Institute of Informatics) 
    doi: \url{https://doi.org/10.20783/DIAS.664}
\end{quote}

Here the dataset's DOI (Digital Object Identifier) does not have version control, and the same DOI (10.20783/DIAS.664) will be used for future releases. This has the advantage of keeping track of the dataset's usage for different versions, but you cannot identify the version of the dataset only from the DOI. For this reason, you need to add the dataset version in the data citation to identify the exact dataset used in the research. 

\section{Representation Learning Using an SSL Framework}
\label{sec-rep}

\subsection{Representation Learning}

Recent advances in self-supervised learning (SSL) demonstrate that neural networks can learn valuable image representations without using labeled data. We based our work on the well-established MoCo v2 framework \cite{moco-v2} for training a feature extractor. Based on contrastive learning, MoCo v2 uses the infoNCE loss function introduced in \cite{cpc-coding} to build a representation space by pulling together positive pairs of images and pushing away negative pairs. Positive pairs are typically created by applying strong augmentations on two different crops of the same image. These augmented images are processed through the feature extractor or a momentum encoder. Both use the same architecture, but the momentum encoder's weights are updated at every training step as a moving average of the feature extractor’s parameters. The two different encodings of the same image are used as positive pairs in the infoNCE loss and the rest of the batch, combined with a queue of passed encoded batches, is used as negative pairs. 

Contrastive learning frameworks require defining two crucial elements: how to identify positive image pairs and which data augmentations to apply for adding random noise to the different views.

First, we discuss positive and negative image pairs. In self-supervised contrastive learning, positive pairs of images are typically defined as two random crops of the same image, assuming that all other images in the batch can be considered different enough to create negative pairs. In our case, we work with typhoon images and, more importantly, sequences of typhoon images. We can include the temporal aspect of a typhoon’s life cycle in the positive pairs selection pipeline. Thus, we chose to establish positive pairs as any pair of different images taken within a 6-hour window of the same sequence.

Second, we discuss data augmentation. It has been shown \cite{simclr} that aggressive data augmentations tend to improve the performances of SSL frameworks. However, the application of data augmentation should be carefully designed for the Digital Typhoon Dataset due to the reasons explained in Section \ref{sec-augmentation}. As a result, we used random solarization, random Gaussian blur, and a 224x224 random crop within a 288x288 center crop (to prevent the eye from always falling in the center).

We trained our feature extractor from scratch for 10,000 training steps, with a batch size of 768. We used the ResNet34 architecture with a cosine learning rate scheduler starting with a value of 0.001. We added a 3-layer MLP projector head during training. This projector head was removed for subsequent experiments, only keeping the ResNet34 architecture with an output dimension of 512. The training was done on a single Nvidia RTX A6000 and took about three days.

\subsection{LSTM Model for Forecasting}
The following experiments aim to evaluate the representation space built with MoCo v2. More specifically, we test the features extracted by our image encoder to see if they can provide a useful signal for forecasting tasks. We used an LSTM (Long Short-Term Memory) model \cite{10.1162/neco.1997.9.8.1735} for time-series modeling of two different forecasting tasks: intensity forecasting (typhoon’s central pressure in hPa in the next 24h) and Extra-Tropical Storm (ETS) transition forecasting.

We used the same simple model architecture for both forecasting tasks, a 3-layer LSTM with a 1024 hidden dimension followed by a 2-layer MLP before prediction. It takes as input a sequence of 3 images, all projected through the frozen image encoder, and outputs a prediction at $t+24$. The three images used as inputs are considered to be the three last known images of the typhoon, taken with a 3-hour interval ($I_{t-6}, I_{t-3}, I_{t}$). All models were trained for 500 epochs with a batch size of 128. The batches were built as follows: 128 sequences are randomly selected, and from each sequence, one sample is randomly selected (3-hour interval inputs + output at $t+24$). It is worth noting that using an LSTM model for this task is unnecessary. LSTMs were chosen because of their known time-series modeling capabilities, but they could be replaced by any other model for the sake of evaluating the potential of the learned representation space of typhoon images.

\begin{figure}[hbt]
    \centering
    \includegraphics[width=1\linewidth]{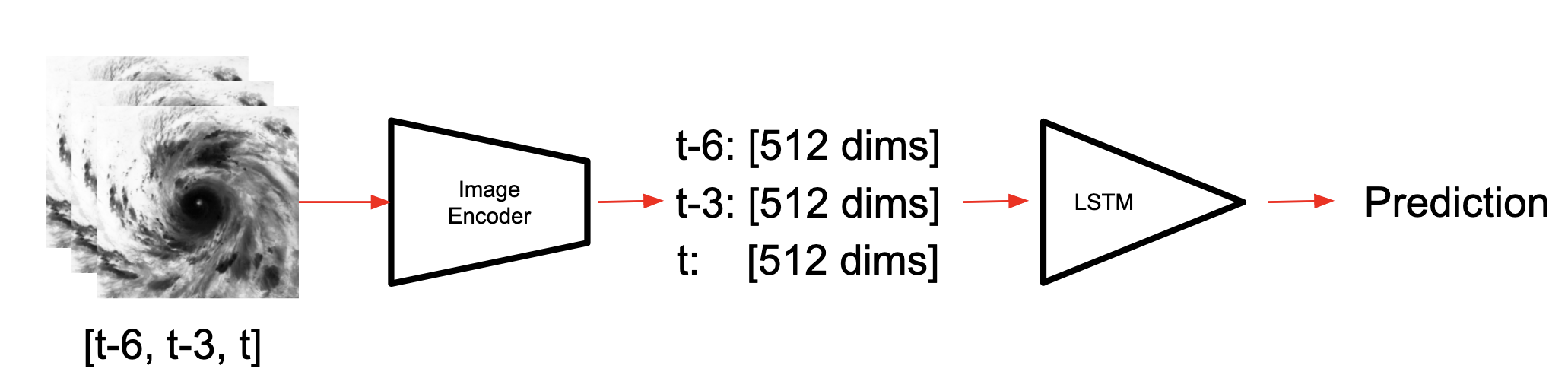}
    \caption{Simplified overview of the forecasting model.}
    \label{fig:enter-label}
\end{figure}

\subsection{Dataset Split}
For intensity forecasting, only images of grade 2-5 (Tropical Cyclones) were used for training and testing whereas for ETS transition forecasting, only sequences containing a transition to a grade 6 storm (ETS) were kept. Using a training/validation/test ratio of 0.7, 0.15, and 0.15, resulted in splits of 652/139/139 sequences for intensity forecasting and 378/79/78 sequences for ETS transition forecasting.

\subsection{Results}
\subsubsection{Typhoon Intensity Forecasting Task}

We kept the same model architecture for different experiments to assess the potential of the learned representations for intensity forecasting. These experiments consisted of changing the signals passed for each of the three input time steps of the LSTM models. Table \ref{tab:pressure} compares the different input configurations we tried. We used the root mean square error (RMSE) in hectopascals (hPa) as an evaluation metric to compare the different models. We also compare the results with two baselines. The first is a naive persistent forecast, which simply repeats the last known value for central pressure. The second baseline has its value taken from \cite{ono_operational_2019}. It represents the forecasting performance of the Typhoon Intensity Forecasting based on the SHIPS method (TIFS) over all Tropical Cyclones in the Northwestern Pacific over the years 2017 and 2018. TIFS is a statistical, regression-based forecasting method aiming to forecast several typhoon parameters, unlike our method, which is specifically trained for central pressure forecasting. This specific row cannot directly be compared with our results since the testing sequences are not the same (139 spread across 40 years for us, compared with 57 spread across 2 years for the TIFS results). We chose to keep it because we believe it gives a good idea of the potential of our methods compared to previous and current ones.

\begin{table}[htb]
    \centering
\caption{RMSE of central pressure (hPa) for 12, 18, and 24 hours forecasting over 139 test sequences.}
\label{tab:pressure}
    \begin{tabular}{|l|c|c|c|} \hline 
         \textbf{Forecast Time}&  \textbf{t+12}&  \textbf{t+18}& \textbf{t+24}\\ \hline 
         Baseline (Persistence forecast)&   9.788&   13.312&  16.249\\ \hline 
         Baseline (TIFS)&  ±9.20&  ±10.80& ±12.00\\ \hline
         Time+Position+Pressure&  5.696&  7.998& 10.043\\ \hline 
         Images only&  8.576&  9.073& 9.982\\ \hline 
         Images + Pressure&  \textbf{5.073}&  \textbf{6.908}& \textbf{8.658}\\ \hline 
         Images + Time + Position + Pressure&  5.240&  7.006& 8.671\\ \hline
    \end{tabular}
\end{table}

\subsubsection{Extra-Tropical Storm Transition Forecasting Task}

Extra-tropical storm transition can happen when a Tropical Cyclone (grades 2-5) transforms into an Extra-tropical storm or ETS (grade 6). Table \ref{tab:ets} shows the results of 3 different models trained using our LSTM pipeline, with image representations and/or metadata as inputs and a binary $t+24$ forecasting as output (1 if the typhoon transitions to ETS, 0 if it stays a Tropical Cyclone). We used three metrics to evaluate these models. We used RMSE to compare the raw outputs of the model with the expert annotation coming from the best track data. After rounding the raw outputs to get a binary output, we used balanced accuracy and F1 score.

\begin{table}[hbt]
    \centering
\caption{Evaluation of different ETS transition forecasting models over 66 test sequences.}
\label{tab:ets}
    \begin{tabular}{|l|c|c|c|} \hline  
         &  RMSE&  Accuracy&  F1\\ \hline  
         Images only&  \textbf{0.177}&  \textbf{0.942}&  \textbf{0.904}\\ \hline  
         Images + Time + Position&  0.180&  0.937&  0.895\\ \hline  
 Time + Position& 0.261& 0.901& 0.862\\ \hline 
    \end{tabular}
\end{table}

\subsection{Discussion}

Table \ref{tab:pressure} shows that the learned representations are effective for central pressure forecasting. The model using only representations of images as inputs is better than both baselines. When comparing the images-only model with the one using only the best track metadata as inputs (time, position, and pressure), the first model performs worse at the $t+12$ and $t+18$ forecasting times but better at the longer $t+24$. The best model, however, is by far when using, as inputs, a combination of the image representations with their corresponding central pressure. We interpret this result as if the model used the provided central pressures as a baseline forecasting and adjusted the final output using the signals provided by the image representations. We also observe that adding more metadata, in our case, position (latitude and longitude) and time (month, day, hour), doesn’t improve the performance of the model. This experiment demonstrates that the representation of typhoon images learned using MoCo contains meaningful information for this specific forecasting task. We also observe that the performance improvement due to using images seems to widen as the forecast time increases. Further experiments and tests at longer forecast times should be conducted to confirm or infirm this observation.

ETS transition forecasting is a very unbalanced task to evaluate with the Digital Typhoon Dataset. Table \ref{tab:ets} shows that using only time and position as inputs lets us obtain a balanced accuracy of 0.901. This high accuracy value is, however, not representative of the actual performances of the model. Qualitative results show that this model performs very badly for forecasting, and this high value is only met because of unbalanced data (typhoons in the dataset spend the major part of their life cycle as a Tropical Cyclone before turning into an ETS). We observe that adding image representations to the model's inputs greatly improves the forecasting performances. This improvement can be seen in the available qualitative results. Again, adding more metadata to the model's inputs doesn't improve its forecasting performance.

\subsection{Future Work}

We showed the potential of using SSL representations of typhoon images for two different forecasting tasks: intensity forecasting and ETS transition forecasting. 
We believe this work only scratches the surface of what can be done by incorporating new computer vision algorithms into meteorology models. We showed that for typhoon images, a self-supervised learning framework like MoCo v2 was enough to extract meaningful features from typhoon images. We leave a lot of room for improvement. Future projects will include incorporating visible image data, testing different image encoder architectures, and evaluating different time-series modeling pipelines. Another avenue to explore is the analysis of the latent space created by the image encoder. Recognizing patterns in typhoon sequences projected in the latent space could eventually improve meteorologists' knowledge of typhoons and their life cycles.

The code for these LSTM experiments and for training the image encoder is available at https://github.com/gafaua/typhoon-latent-forecast.

\section{Typhoon Center Estimation Task} 
\label{CET}

These typhoon images also enable the completion of other challenging tasks. One of them is estimating the typhoon center. This task is especially challenging when a typhoon is weak, and the eye is not clearly visible. In this case, the cloud patterns do not show characteristic signs, and even domain experts fail to identify the typhoon center. Hence, we develop a machine learning model that automatically estimates the typhoon center. 

\subsection{Method}

We can formulate this problem as an object detection task. A wide range of methods is available for object detection. From R-CNN \cite{rcnn} (Region-based Convolutional Neural Networks), YOLO \cite{yolo} (You Only Look Once) to the more recent SAM \cite{sam} (Segment Anything Model). All these models have their specificity, but almost all rely on bounding boxes for the training phase. In our case, however, we only need the location of the typhoon center, not its size. To avoid using unnecessary information for training, we focus on models that do not require bounding boxes. 

We thus choose to adopt the model introduced by Javier Ribera et al. \cite{lowbb}. This method introduces the weighted Hausdorff distance as a loss function for a U-Net \cite{unet} architecture. The architecture is an encoder-decoder model acting as a feature extractor. It is then coupled with the loss function previously mentioned. This loss was chosen for its capacity to grasp an object, even small ones. This model takes only satellite images as input and produces a heatmap as an output. The heatmap can be seen as a probability map, and we
apply a deterministic decision method on the probability map to estimate the center. In Javier Ribera et al., they used the Gaussian mixture to detect multiple points. In our case, however, we know that only one point, namely the typhoon center, is contained in the image, so we select the maximum of the heatmap as the estimated center of the typhoon.

\begin{figure}[ht]
    \centering
    \includegraphics[width=1\linewidth]{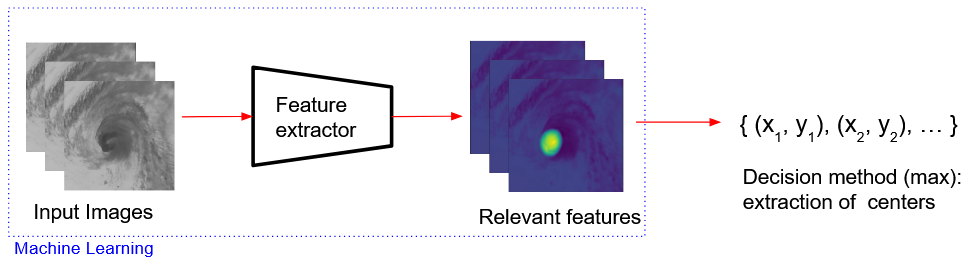}
    \caption{Simplified overview of the center estimation pipeline.}
    \label{fig:overview_cet}
\end{figure}

\subsection{Dataset for Typhoon Center Estimation}

Here we create a new dataset for this task from the original Digital Typhoon Dataset. First, we filter the dataset to select data with the grade from 2 to 5. Second, we introduce a randomness factor for our input data. Knowing that each image is centered on the eye of the typhoon, we randomly crop a portion of the image, including the typhoon center. This portion is chosen to be 256 by 256 pixels, about 1250 km wide. Moreover, because the eye of the typhoon can be located close to the edge of the image, we introduce $\epsilon$, which is the minimum distance (in pixels) between the eye and any of the edges. This parameter preserves the context around the eye to provide enough information for the model training. Our dataset is then composed of cropped images and the eye coordinates after the cropping transformation. Note that we compute the new coordinates knowing that this operation is similar to a translation of the eye on the image. These coordinates are then normalized between 0 and 1 as float values.

As stated in Section \ref{sec-augmentation}, we must carefully select data augmentation techniques in our experiments. To check the effect of image rotation augmentation, we implemented another step in the cropping operation. With a probability $p$, a typhoon image is rotated around its eye by an angle $\theta$. This angle is also chosen randomly between the interval $[\theta_{\min}; \theta_{\max}]$ fixed before the training phase. This interval must be small so that the spatial deformation of the typhoon does not distort the meaning of the image.  

Using the newly created dataset, we split it using a training/validation/test ratio of 0.7, 0.15, and 0.15. The dataset is shuffled and time dependencies are discarded, which means that the sequence structure of the typhoon is not preserved.

\subsection{Results}

\begin{figure}[ht]
    \centering
    \includegraphics[width=0.5\linewidth]{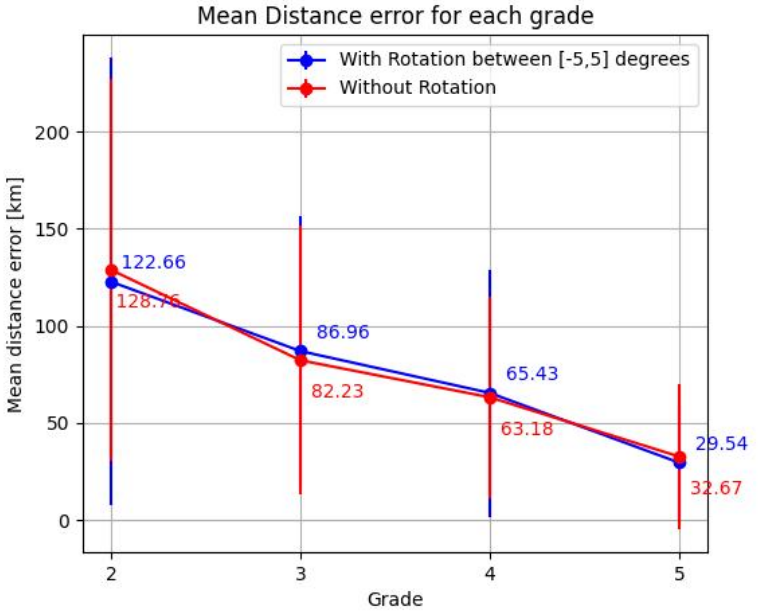}
    \caption{Mean distance error for the two experiments regarding the typhoon's grades.}
    \label{fig:erwan_fig1}
\end{figure}

We trained two models with the architecture as previously presented. Both models used the cropping operation. Moreover, $\epsilon$ was arbitrarily set to $\epsilon = 30$ pixels for the two models. One of the models was trained with rotation augmentation with a probability $p=0.5$ and with $\theta$ in degrees randomly set for a value between -5 and 5 degrees. The models were trained for 140 epochs. Figure \ref{fig:erwan_fig1} presents the result of this experiment.  The mean distance error in kilometers is plotted for each grade. This distance is the Euclidean distance between the estimated center and the ground truth. In our case, because of the resolution of the image, one pixel is equal to around 5 km. Two curves are plotted, referring to the two different experiments. The blue curve shows the results of the experiment with rotation augmentation, and the red curve without it. Both curves are coupled with their standard deviation displayed by the vertical lines.

\subsection{Discussion}
Figure \ref{fig:erwan_fig1} shows that it is relatively effective on all the grades. However, we observe a linear relationship between error and typhoon intensity, where a strong typhoon has a smaller error. This is explained by the fact that a stronger typhoon has a clearer eye around the center. On the other hand, for weaker typhoons, it is still a challenging task when cloud patterns are unclear. Moreover, it seems that rotation augmentation does not improve the performance. Meteorologists pointed out that even small rotation angles may influence the direction of the wind and cloud, thus creating a bias in the dataset.
Qualitatively, the size of the heatmap spot is larger for weaker typhoons than for stronger ones, suggesting that uncertainty increases for weaker typhoons. Figure \ref{fig:erwan_fig1} also proves this point when we look at the standard deviation.  
We also found cases where two maxima were separated. In this case, our method only chooses a peak with the absolute maximum, potentially missing other peaks that may correspond to the actual center. Hence, we suggest that human intervention may be required to deal with this type of uncertainty. We also observe that $\epsilon$ is a critical parameter because it reduces the image by discarding a part of the image that may be useful for center estimation. Regarding the input image size, the model performs worse for images with a bigger size.

There are many improvements to enhance the performance. Working with an entire satellite image requires the improvement of scalability in the machine learning model. Finding another data augmentation satisfying the conditions mentioned above is of interest. It could be used for broad tasks based on typhoon images. Using another model, such as the CenterNet \cite{centernet} architecture, could also be a research topic.  

 \section{Model Generality Across Basins and Hemispheres}
\label{sec-basin}

This task focuses on the behavior of a machine learning model applied to different basins of typhoons to check if machine learning models can work universally across all basins. Thus, we focus on studying the performance of different tasks applied to two different basins: the WP dataset and the AU dataset. We will address three tasks: the center estimation task, the intensity  classification task, and the intensity regression task.  
 
 \subsection{Method}
The models are trained on the new dataset for all three tasks. We use the split of the WP dataset to train all models. The aim is to test all the models trained on the WP dataset and study the behavior on the AU dataset. We introduce a horizontal flip to align the direction of circulation between the two datasets.

\subsubsection{Center Estimation Task}
This task is identical to the previously presented task in Section \ref{CET}. We take the same model without rotation augmentation. This model was trained on WP images only. To demonstrate the necessity of the horizontal flip, we test our model on AU images that are not horizontally flipped.

\subsubsection{Intensity Classification Task}
This task aims to predict the typhoon's grade at a time $t$ only using the image. Here again, only the grade 2 to 5 typhoons are used. Moreover, the images are 384x384 pixels wide which is the maximum size to allow efficient training with our hardware. Regarding the data splitting, we used a splitting just like in Section \ref{CET}. We used two different architectures, ResNet \cite{resnet} and Vision Transformer (ViT) \cite{vit}, to perform feature extraction. We then compared the results for both architectures to determine which performs better for this task. We used the Pytorch implementation of both models. We used the \texttt{resnet34} and the \texttt{vit\_b\_16} size model from the torch library.  The loss is the cross entropy loss. The models are trained for 60 epochs with a learning rate of $5 \cdot 10^{-3}$.

\subsubsection{Intensity Regression Task}

This task aims to go a step further than the previous one. We aim to estimate central pressure using only typhoon images. However, the data splitting must be different. Here, the data is sorted by sequences to prevent the model from performing an interpolation task rather than a regression one.
But for this section, we only use ViT \cite{vit} to perform this task. The model is the same; only the loss is changed to the L1 loss. The model is trained for 200 epochs with a learning rate of $5 \cdot 10^{-3}$.

\subsection{Results}

\subsubsection{Center Estimation Task}
\begin{figure}[ht]
    \centering
    \includegraphics[width=0.5\linewidth]{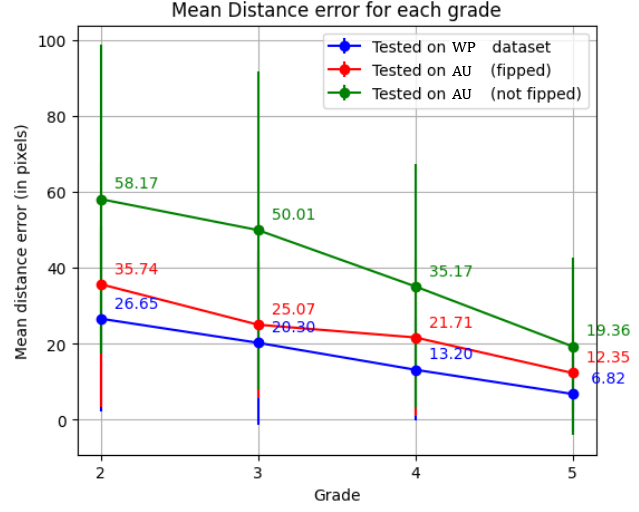}
    \caption{Mean distance error for the center estimation task with the two test basins.}
    \label{fig:gen_cet}
\end{figure}

Figure \ref{fig:gen_cet} shows the mean distance error of center estimation, just like Figure \ref{fig:erwan_fig1}.  Here, we show the performance of the WP and AU test sets. Moreover, we also display the performance of the test on the AU test set but without a horizontal flip. Here, the distances are no longer in km but in pixels. 
\subsubsection{Intensity Classification Task}

To evaluate the classification performance, we use the precision metric. This metric shows the proportion of positive predictions. It is defined as follows:
\begin{equation}
    \text{precision} = \frac{\text{TP}}{\text{TP} + \text{FP}}
\end{equation}
where TP is True positive, and FP is False positive. Table \ref{tab:tab_precision_dataset} shows the precision for both models on two basins regarding the typhoon's grade. Blue has the lowest value, and red has the highest.

\begin{table}[ht]
 \centering
\caption{Model precision on two basins regarding the typhoon grades.}
\label{tab:tab_precision_dataset}
\begin{tabular}{c|ll|ll|}
\cline{2-5}
\multicolumn{1}{l|}{}        & \multicolumn{2}{c|}{ViT}                                                         & \multicolumn{2}{c|}{ResNet}                                                      \\ \hline
\multicolumn{1}{|l|}{Grades} & \multicolumn{1}{l|}{WP}                          & AU               & \multicolumn{1}{l|}{WP}                          & AU                          \\ \hline
\multicolumn{1}{|c|}{2}      & \multicolumn{1}{l|}{0.856}                        & 0.670                        & \multicolumn{1}{l|}{0.912}                        & {\color[HTML]{FE0000} 0.738} \\ \hline
\multicolumn{1}{|c|}{3}      & \multicolumn{1}{l|}{0.773}                        & 0.31                         & \multicolumn{1}{r|}{0.837}                        & 0.343                        \\ \hline
\multicolumn{1}{|c|}{4}      & \multicolumn{1}{l|}{{\color[HTML]{6434FC} 0.744}} & {\color[HTML]{6434FC} 0.28}  & \multicolumn{1}{l|}{{\color[HTML]{6434FC} 0.825}} & {\color[HTML]{6434FC} 0.305} \\ \hline
\multicolumn{1}{|c|}{5}      & \multicolumn{1}{l|}{{\color[HTML]{FE0000} 0.927}} & {\color[HTML]{FE0000} 0.785} & \multicolumn{1}{l|}{{\color[HTML]{FE0000} 0.923}} & 0.698                        \\ \hline
\end{tabular}
\end{table}

\subsubsection{Intensity Regression Task}
\begin{figure}[ht]
    \centering
    \includegraphics[width=0.5\linewidth]{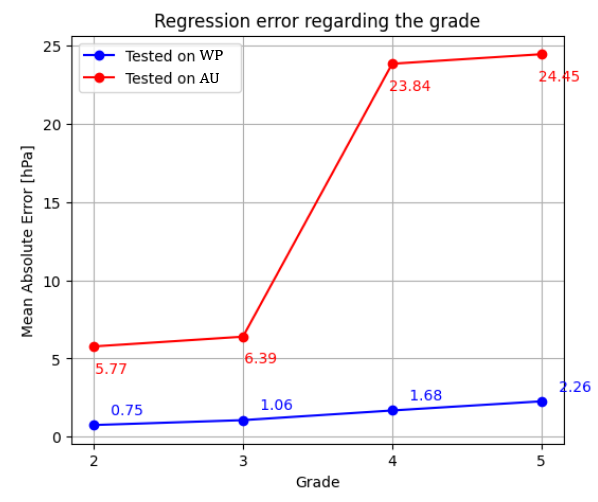}
    \caption{Regression error for the two test basins using ViT.}
    \label{fig:press_regression}
\end{figure}

Figure \ref{fig:press_regression} shows the estimation error of the pressure for grades 2 to 5. The pressure error (in hPa) is the absolute difference between the estimation and the ground truth. We display the results for the two basins.

\subsection{Discussion}

In Figure \ref{fig:gen_cet}, the gap between the red and green curves throughout all grades indicates that the horizontal flip is necessary. This flipping increases homogeneity (hence comparability) in the training and test data, thus between the two basins. 
Figure  \ref{fig:gen_cet},  Figure \ref{fig:press_regression} and Table \ref{tab:tab_precision_dataset} show a gap in performance between the two test basins for all three tasks. This suggests that just flipping the image may not be enough to homogenize images from the two basins. 
 Moreover, Table \ref{tab:tab_precision_dataset}, showing a gap in the values, may have a different interpretation. This classification task heavily relies on labels, but labels were given by two different agencies. Knapp and Kruk \cite{agencies_diff} pointed out that there may be interagency differences.  Especially the gap in performances for grades 3 and 4 in Table \ref{tab:tab_precision_dataset} could be due to the different transition rules between grades. This observation is also made using the confusion matrix. It could also explain why the two models have similar behavior. Considering the human factor, the performance may indicate a data problem instead of an architecture problem. 
 
\section{Conclusion}

We have introduced the Digital Typhoon Dataset V2 for machine learning and meteorology communities to promote data-driven research on tropical cyclones. This is the first update of the dataset since November 2023, and we significantly updated the dataset to the spatial dimensions by adding data from the southern hemisphere. In future versions of the dataset, we are planning to add more dimensions to the dataset. One direction is adding more channels in Infrared, Water Vapor, and Visible. Another direction is to increase the temporal frequency, such as 30 minutes, 10 minutes, and even 2.5 minutes. 

In these extensions, we need to shorten the period of the dataset, because other channels and higher frequency data are available only for newer satellites. However, those new datasets will open up the possibility of new tasks, allow us to ask new research questions, and lead to new scientific discoveries and effective solutions to societal and sustainability issues.

\begin{ack}
Two of the authors, Erwan Dzik and Gaspar Faure, have been supported by the international internship program of the National Institute of Informatics.
\end{ack}

\bibliographystyle{plain}
\bibliography{references}

\end{document}